# External Steering of Vine Robots via Magnetic Actuation

Nam Gyun Kim*, Nikita J. Greenidge*, Joshua Davy, Shinwoo Park, James H. Chandler, Jee-Hwan Ryu, Pietro Valdastri

[1] Interactive Robotic Systems (IRiS) Lab, Department of Civil and Environmental Engineering, Korea Advanced Institute of Science and Technology, Daejeon, Republic of Korea

[2] Science and Technologies Of Robotics in Medicine (STORM) Lab, School of Electronic and Electrical Engineering, University of Leeds, Leeds, UK

*Corresponding Author: Nikita J. Greenidge, elnjg@leeds.ac.uk*

## ABSTRACT

This paper explores the concept of external magnetic control for vine robots to enable their high curvature steering and navigation for use in endoluminal applications. Vine robots, inspired by natural growth and locomotion strategies, present unique shape adaptation capabilities that allow passive deformation around obstacles. However, without additional steering mechanisms, they lack the ability to actively select the desired direction of growth. The principles of magnetically steered growing robots are discussed, and experimental results showcase the effectiveness of the proposed magnetic actuation approach. We present a 25 mm diameter vine robot with integrated magnetic tip capsule, including 6 Degrees of Freedom (DOF) localization and camera and demonstrate a minimum bending radius of 3.85 cm with an internal pressure of 30 kPa. Furthermore, we evaluate the robot's ability to form tight curvature through complex navigation tasks, with magnetic actuation allowing for extended free-space navigation without buckling. The suspension of the magnetic tip was also validated using the 6 DOF localization system to ensure that the shear-free nature of vine robots was preserved. Additionally, by exploiting the magnetic wrench at the tip, we showcase preliminary results of vine retraction. The findings contribute to the development of controllable vine robots for endoluminal applications, providing high tip force and shear-free navigation.

## INTRODUCTION

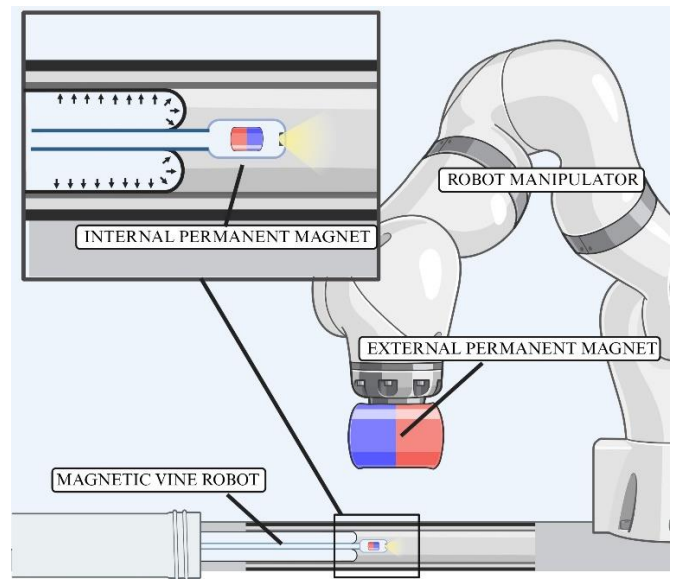

*Figure 1: Overview of the proposed system. The robot consists of a growing section with a tip containing an Internal Permanent Magnet (IPM). By manipulation of the magnetic field via a large External Permanent Magnet (EPM) mounted to a robotic manipulator, the growing robot can be steered without complex internal mechanisms.*

Inspired by the growth and locomotion strategies of climbing plants observed in nature[1,2], vine robots possess a unique ability to adapt their shape, to passively deform around obstacles and navigate complex environments, making them highly versatile for various applications. Their utility has been considered for application to medical procedures[3–5], exploration[6], and environmental monitoring[7].

Vine robot growth is typically driven by pneumatic actuation, where pressurization of an inverted inflatable cylinder causes internal material to be continuously transported and inflated at the tip, resulting in lengthening of the robot's body. This growth mechanism (extension from the tip), as opposed to translational motion, mitigates friction with the surrounding environment, therefore allowing the robot to move without drag. Furthermore, this mechanism for growth enables the robot to adapt to existing channels, facilitating gentle navigation within the environment[7]. However, vine robots operating in open space and without any additional steering mechanism lack the ability to control tip position and orientation, and thus growth direction. Therefore, the development of steering mechanisms to facilitate this capability while preserving the unique properties of the vine robot represents a significant research challenge.

To address this issue, conventional approaches to steering vine robots have relied on embedding actuation systems throughout their bodies, such as Series Pneumatic Artificial Muscles (SPAMs)[8,9], Fabric Pneumatic Artificial Muscles (fPAMs)[10], and tendon-driven actuation[11]. Fluidic approaches typically offer moderate distributed actuation along the length of the robot's body, realizing integrated designs with limited curvatures. Conversely, tendon-driven approaches can typically achieve higher curvatures, however, suffer from buckling and challenges due to tendon friction when the robot encounters consecutive bends.

As an alternative to whole-body actuation methods, tip-focused steering has been explored using rigid mechanisms that steer only the tip of the robot. This approach enables the robot to achieve higher curvature than the conventional whole-body steering, which is particularly useful for operating within small spaces. For example, a vine with an additional internal pneumatic bending mechanism was used to navigate a vine robot in a confined space[12], a hybrid vine robot was developed to induce buckling via an internal rigid steering mechanism to achieve high bending angles at discrete points along its length[13], and a rolling contact-based tip steering mechanism was proposed for improved target reachability[14].

Until now, the forces that trigger the steering of vine robots have been grounded within the robot itself, requiring the integration of additional mechanisms. Moreover, mounting and maintaining components at the robot's tip represents a unique challenge for vine robots due to their everting nature, typically necessitating either a rigid cap-like mount[15], internal pneumatic mechanisms[16] or a complicated control process[17]. Achieving tip-steering or stable tip mounts in vine robots has therefore come at the cost of having to include mechanisms that occupy significant space within the vine, interfere with their compliant nature, limit miniaturization potential and increasing the overall mechanical complexity of the system. In this paper, we investigate for the first time the concept of external actuation of the tip of a vine robot through magnetically induced torques and forces.

By integrating a magnetic element into the structure of the robot and manipulating the external field, forces and torques can be induced on the robot's body. This principle has been proposed for use in medical applications such as tethered endoscopic navigation of the gastrointestinal (GI) tract[18], and in soft magnetic catheters for navigation through the cardiovascular system[19] or the bronchial tree[20]. Conventional flexible endoscopes and catheters are all introduced into the human body by pushing them from the distal side. In order for them to advance inside a convoluted anatomy, they all have to apply pressure on the lumen wall whenever they approach a bend. This often leads to complications for the patient such as pain or perforations. Magnetic actuation facilitates designs that can conform better to the anatomy and can be deployed under open and closed loop control strategies using various localization methods. However, these methods suffer limitations due to their low actuation force and the build-up of friction during insertion[21].

The generation of actuating magnetic fields is possible via systems of electromagnetic coils[22] or the use of manipulated External Permanent Magnets (EPMs)[23,24]. The use of manipulated EPMs comes with the advantage of larger workspace, lower power requirements, but increased control complexity[23]. Relatively, systems based on electromagnets produce weaker fields for the same size system and require large external power supplies and cooling systems. However, control is simplified due to the linear relationship between produced field and applied current.

In this study, we explore the application of an EPM mounted to a 7 Degree of Freedom (DOF) robot manipulator to generate the magnetic fields required for tip steering a magnetic vine robot (MVR) (See Figure 1), building on our previous work on magnetically manipulated flexible endoscopes with 6 DOF real-time tip localization[18]. By introducing a novel mounting system, we allow for the secure placement of a magnetic element at the robot's tip during eversion, enabling the robotically controlled EPM to induce the desired forces and torques required for precise tip navigation. This MVR approach aims to overcome the limitations of current vine robot steering techniques and allow the compliant, shear-free growing nature of vine robots to be harnessed for improved navigation in delicate environments like endoluminal anatomies (e.g. the gastrointestinal tract, the bronchial tree or the cardiovascular system).

We investigate the capability of magnetic actuation to facilitate high-curvature, tip-focused steering and independent tip orientation of vine robots without relying on complex internal mechanisms. Further experimentation evaluates how the MVR's growing nature can effectively expand the workspace of magnetic endoluminal navigation with its enhanced axial pushing force and eliminate tether drag. Furthermore, we validate the reproducibility and repeatability of the control of the MVR, by studying deviation across multiple fixed trajectories. We also present magnetic tip suspension tests that demonstrate the MVR's maintained shear free motion despite the integration of a rigid component at the tip and a preliminary examination of retraction.

While in this work the magnetic field is generated by a robotic EPM, the proposed approach can be generalized to any method of magnetic actuation, provided sufficient magnetic field can be generated and controlled.

## MATERIALS AND METHODS

### Principles of Magnetic Vine Robots

In this section, we explain the principles of MVRs. In particular, we discuss the mathematical fundamentals of magnetic manipulation in order to justify the design choices in the MVR hardware as-well as experimental setup. We also discuss the considerations of the vine portion and how this pressurized body affects the overall deformation of the robot. For simplicity, internal forces and gravity are omitted. We assume that the bending stiffness of the tether is negligible compared to that of the inflated vine, and we consider magnetic forces to act directly on the robot's tip.

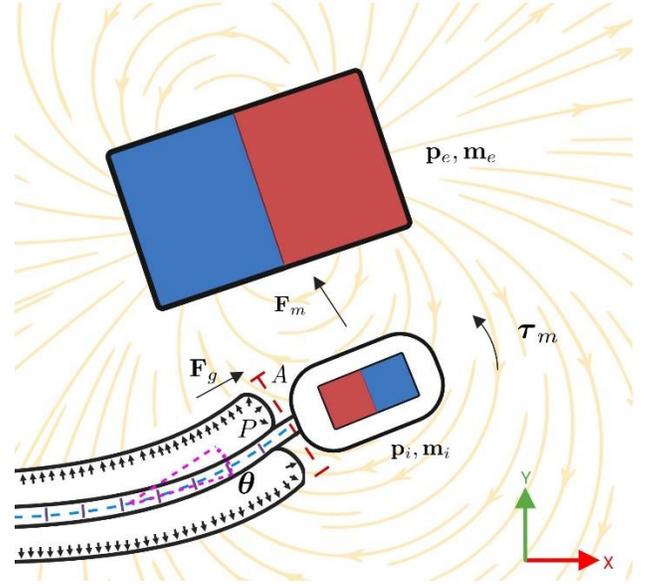

*Figure 2: Diagram of contributing forces and torques to the movement of the robot. The total force is the sum of the pressurized growing section of the robot and the attractive force between EPM and IPM. The bending moment is a balance between the magnetic wrench of the IPM within the EPM's field and the restoration moment of the growing section. θ is the deflection of the vine body.*

The use of both magnetic and growing actuation compounds the large tip forward force provided by the growing system with the magnetic forces and torques (wrench) on a magnet within an external field[7,24]. This combination of actuation methodologies enables the remote steering of MVRs though external magnetic field manipulation.

To realize this combined actuation method, we propose the introduction of an Internal Permanent Magnet (IPM) at the tip of a pneumatically actuated vine robot. When operated within an external magnetic field, the IPM and thus the tip of the vine robot experiences magnetic forces and torques which can be used to steer the vine robot. To provide suitable external magnetic fields for positioning and steering, we manipulate the pose of an External Permanent Magnet (EPM) around the vine robot's workspace. (See Figure 2).

The magnetic field **B** at a point **r** produced by an EPM with magnetic moment $\mathbf{m}_e$ is given as

**Equation 1**

$$\boldsymbol{B}(\boldsymbol{m}_e, \boldsymbol{r}) = \left(\frac{\mu_0}{4\pi\|\boldsymbol{r}\|^3}(3\hat{\boldsymbol{r}}\hat{\boldsymbol{r}}^T - I)\right)\boldsymbol{m}_e,$$

where we consider $\mathbf{r} = \mathbf{p}_e - \mathbf{p}_i$, the relative displacement between the EPM position $\mathbf{p}_e$ and IPM position $\mathbf{p}_i$. $\hat{\mathbf{r}}$ is the direction vector of $\mathbf{r}$, $\hat{\mathbf{r}} = \frac{\mathbf{r}}{|\mathbf{r}|}$. I is the identity matrix and $\mu_0$ is the vacuum permeability equal to $4\pi \times 10^{-7}$ Hm$^{-1}$.

The magnetic torque $\boldsymbol{\tau}_m$ on the IPM is

**Equation 2**

$$\boldsymbol{\tau}_m = \mathbf{m}_i \times \mathbf{B}.$$

$\mathbf{m}_i$ is the magnetic moment of the IPM equal to

**Equation 3**

$$\mathbf{m}_i = \frac{\mathbf{B}_r v}{\mu_0},$$

where $\mathbf{B}_r$ is the residual magnetic flux density vector, $v$ is the volume of magnetic material[23].

The spatial gradient of the EPM field is given as

**Equation 4**

$$\mathbf{B}_\nabla(\mathbf{m}_e, \mathbf{r}) = \frac{3\mu_0}{4\pi \|\mathbf{r}\|^4}(\mathbf{m}_e \hat{\mathbf{r}}^T + \hat{\mathbf{r}} \mathbf{m}_e^T + \hat{\mathbf{r}}^T \mathbf{m}_e(\mathbf{I} - 5\hat{\mathbf{r}}\hat{\mathbf{r}}^T)),$$

which relates to the magnetic force on the IPM $\mathbf{F}_m$ as

**Equation 5**

$$\mathbf{F}_m = \mathbf{B}_\nabla^T \mathbf{m}_i.$$

The growing portion of the robot exerts a force on the tip due to the pressurized section of the robot. The pushing force under quasi-static conditions can be simplified:

**Equation 6**

$$F_g = \frac{1}{2} PA - C,$$

where $P$ is the internal pressure of the vine robot, $A$ is the cross-sectional area of the vine robot, and $C$ is a drag term related to material deformation, tip velocity etc. Details can be found in the work of Blumenschein et al[25].

The inflated vine body, creates a restoration moment $\boldsymbol{\tau}_g$ which is a non-linear function of the diameter of the vine $D$, deflection angle $\boldsymbol{\theta}$ and $P$ the internal pressure $\boldsymbol{\tau}_g = f(\boldsymbol{\theta}, P, D)$. Other smaller factors that affect this restoration moment include vine material elasticity and tether tension.

The wrench of the MVR will therefore depend on the pose of the robot, the diameter of the vine, the applied pressure and the relative EPM-IPM pose. From Equations 1-5, it can be noted how magnetic torques and magnetic forces drop off with distance (i.e. $\frac{1}{r^3}$ and $\frac{1}{r^4}$, respectively). The overall tip force is the summation of the force contributions of the pressurized vine body and the magnetic force on the IPM (Equations 5 and 6). In order to steer the vine body, the overall magnetic wrench applied must overcome the restoration moment introduced by the pressurized vine body. An IPM must therefore be selected with sufficient volume to generate the required wrench for the specific external magnetic actuation system used while fitting within the constraints of the overall MVR diameter.

## Hardware of Magnetic Vine Robots

To achieve magnetic steering of vine robots, three main components are required. A magnetic tip, a system to generate and control an external magnetic field and a vine robot that allows for tip locking and tether feeding.

### Magnetic Tip

The magnetic tip was constructed from a 20 mm x 35 mm 3D printed cylindrical clear resin shell (Formlabs) containing an 11 mm x 22 mm axially magnetized cylindrical NdFeB N52 permanent magnet (K&J Magnetics) wrapped in a flexible magnetic sensor array circuit for 6 DOF real-time localization (±2 mm position and ±3° orientation accuracy)[18]. An endoscopic camera and LED were included to provide visual feedback from the tip, as well as a tool channel for passing tools down the tether (See Figure 3).

### Magnetic Field Generation

For external magnetic field manipulation, a 101 mm x 101 mm NdFeB N52 EPM (Magnetworld AG) with a localization coil was mounted as the end effector of a 7-DOF serial robotic manipulator (14-kg payload, LBR, KUKA). Joint angles of the arm were controlled via a Cartesian joystick controller using ROS[21].

### Vine Robot

The vine, and vine base were designed to allow the magnetic tip's sensor cable tether to be transported by utilizing a scrunched material stacking method, similar to the origami-inspired base developed by Kim et al.[17]. The notable improvements in our design include the integration of movable rollers, a pressure separator that facilitates the free transport of high gauge sensor wire into the pressurized base, and a sealing ring that continuously synchronizes internal pressure while locking the robot tip and facilitating tether feeding.

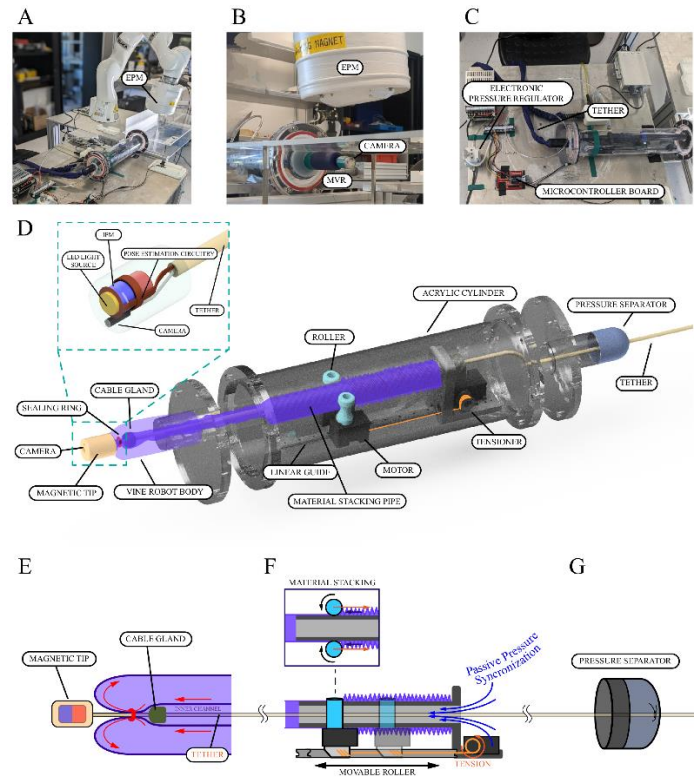

*Figure 3: Hardware of the Magnetic Vine Robot (MVR). (A) Overview of the MVR system. (B) Vine robot with magnetic tip. (C) Electrical components. (D) CAD model of the MVR. (E) Schematic of the growing tip. (F) Schematic of the material stacking mechanism. (G) Schematic of the pressure separator.*

As shown in Figure 3, the movable rollers are placed on a linear guide and pulled tight by a secondary tensioning motor. With sufficient tension pulling the linear guide, the material can be kept in a fully scrunched state. The fully stacked material just behind the roller then generates a pushing force, which automatically pushes the rollers forward during the stacking of the material. This configuration facilitates even vine material stacking across the stacking pipe allowing more material to be stacked in the same length of pipe without getting stuck just behind the roller.

The entire stacking system is placed in an acrylic cylinder with openings on both sides to pressurize the system. Placing an elastic sealing ring at the tip prevents air leakage and therefore continuously synchronizes the inner pressure with outer pressure for inner channel security.

The vine base isolates the pressure inside the cylinder from the atmospheric pressure, while the high gauge cable can still freely move through via the pressure separator. The pressure separator is made of ripstop nylon to minimize friction between the separator and the tether, tightening the tether by elastic rubber built in the fabric.

By placing the sealing ring at the front tip and maintaining equal pressure levels between the inner channel and the main body, compression and, therefore, friction can be relieved to allow the wires to move freely inside the inner channel of the vine. The elastic sealing ring behind the magnetic tip allows the position of the magnetic tip to be maintained, preventing the spitting of the tip during growing. A cable gland prevents this elastic sealing ring from sliding down the tether during growing (See Figure 3E).

Figure 4 describes the fabrication of the inflatable vine robot body. The cylindrical shape was made by sewing Ripstop nylon fabric (Seattle Fabrics) and sealing it with silicone adhesive. A resulting diameter of 25 mm was achieved. The vine body's diameter is deliberately larger than that of the magnetic tip to ensure that the frictionless navigation aspect of vine locomotion can be retained.

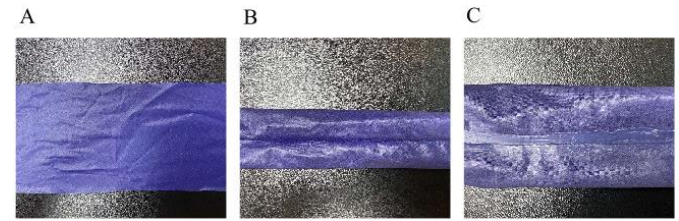

*Figure 4: Fabrication of the growing body from Ripstop nylon fabric. (A) Fabric weave oriented at 45°. (B) Seam sewn. (C) Seam sealed with silicone adhesive.*

To reduce the restoring moment that opposes magnetic actuation while the MVR tip is bending, we placed the fabric at a 45-degree angle when sewing it into a cylindrical shape. This orientation leverages the anisotropic stretchability of ripstop nylon, aligning it with the vine robot's principal axis and minimizing volume changes during curvature, thereby reducing the restoring moment[10]. Finally, to control the growth of the vine robot, the pressure was regulated using an electronic pressure regulator (SMC ITV-1010) and the rollers and tensioner were controlled using motors (Robotis Dynamixel XL330-M288-T). Both the motors and the electronic pressure regulator were controlled using a microcontroller (Robotis OpenRB-150).

# EXPERIMENTAL RESULTS

## Tip Force Measurements

The resultant force at the tip of the MVR is a combination of the magnetic force between IPM and EPM and the growing force of the pressurized section. The magnetic force $\mathbf{F}_m$ is a function of the relative distance and orientation between IPM and EPM (Equation 4 and Equation 5), while the growing force $\mathbf{F}_g$ is a function of the applied pressure (Equation 6). To evaluate the relative contribution of each force, we varied the IPM-EPM separation distance at a range of static growing pressures. Figure 5A shows the experimental setup with the MVR constrained within a tube with an inner diameter matching the diameter of the MVR. The tip was placed against a force sensor (Nano 17, ATI Industrial Automation), and the EPM positioned above. Prior to each test, the EPM was positioned at a starting height of 100 mm and a lateral distance of 85 mm relative to the MVR's tip magnet. The height of 100 mm was chosen as the lower limit of the clinically applicable range, informed by previous work showing an effective actuation range of 100 mm to 150 mm for applications such as colonoscopy[21]. The lateral distance of 85 mm was experimentally found through stepwise measurements to maximize tip force at the chosen height. The pressure was set to a fixed value for each test ranging from 0-30 kPa in steps of 5 kPa.

At each pressure, the EPM height d was increased in steps of 50 mm up to a maximum height of 300 mm. At each height, the force on the load cell was observed over a period of 20 seconds and the mean value was recorded as the pushing force.

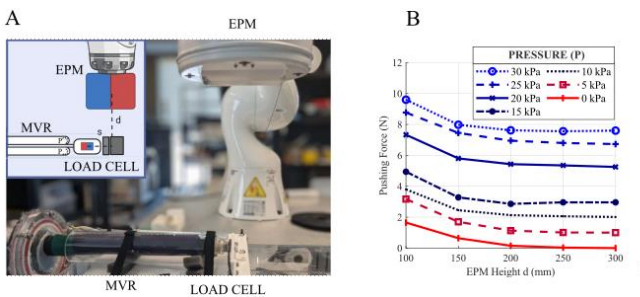

*Figure 5: Tip force measurements. (A) Arrangement of MVR in the perspex tube with the load cell. (B) Graph of pushing force of MVR with varying pressure P and EPM height d.*

Figure 5B shows the average pushing force results as a function of EPM height at each pressure. It can be observed that pushing force converges with greater height of the EPM. The plateau reached at each pressure represents the respective contribution of growing force to the total pushing force. At higher pressures, the pushing force is dominated by the contribution of the growing portion. For example at 30 kPa, the growing force is 79 % of the overall force with an EPM height of 100 mm. As expected, at lower pressures the magnetic forces form a larger contribution, however, when EPM height is greater than 200 mm, the contribution is negligible.

## Steering of MVR

In order to steer the MVR using magnetic actuation, the bending moment from the magnetic wrench must overcome the restoration moment of the growing section. To demonstrate this capability, we consider the maximal bending of the MVR that is achievable in free space using magnetic actuation.

The MVR was positioned in free space under a constraining plate to prevent contact between the EPM and the tip (See Figure 3B) and grown through an acrylic tube of similar diameter to constrain the proximal portion and allow control over its unconstrained grown length. The MVR was pressurized to 30 kPa and the separation between EPM and IPM height was set at 160 mm. As observed in the results of the tip force measurements, at this distance and pressure, the magnetic force contribution in the direction of travel is minimal. The MVR was subsequently grown until its unconstrained length, the length from the end of acrylic tube to the MVR tip, was 10 cm (Figure 6A) and 20 cm (Figure 6B) respectively. The EPM was then manipulated to achieve the minimum bending configuration of the MVR body without buckling. It can be observed that the MVR achieves a minimal bending radius of 4.95 and 3.85 cm in the 10 cm and 20 cm cases, respectively. It can also be observed how the MVR tip can still be steered independently of the bending of the growing section by rotation of the EPM, as indicated by the red arrows in Figure 6A and B. To show the ability of the MVR to achieve high (non-constant) curvature, the EPM was manipulated so that the MVR was bent beyond the point of buckling as shown in Figure 6C (See Supplementary Video 1) where the proximal constraint acts as a pivot point for the vine's body.

To study the repeatability of the bending control of the MVR using the EPM, we considered fixed circular trajectories of the EPM. Here, the EPM was positioned at a height of 160 mm and moved in a fixed circular trajectory with a radius equivalent to the MVR's length from 0 to 130 degrees. These tests were conducted over a range of pressures (P = 10.0 kPa, 20.0 kPa, 30.0 kPa) and robot lengths (l = 100 mm, 150 mm, 200 mm) with five repeats each. Starting from a straight pose, as the MVR is bent, it forms an approximate constant curvature which was quantified as a mean bending radius via image analysis (see supplementary materials). Table 1 summarizes these results. It can be observed that bending radii is approximately constant and independent of applied pressure and robot length. This shows the dominance of the magnetic wrench over restoration torque within the vine body and the repeatability of the actuation methodology to achieve high curvature bending of the MVR (See Figure 6D).

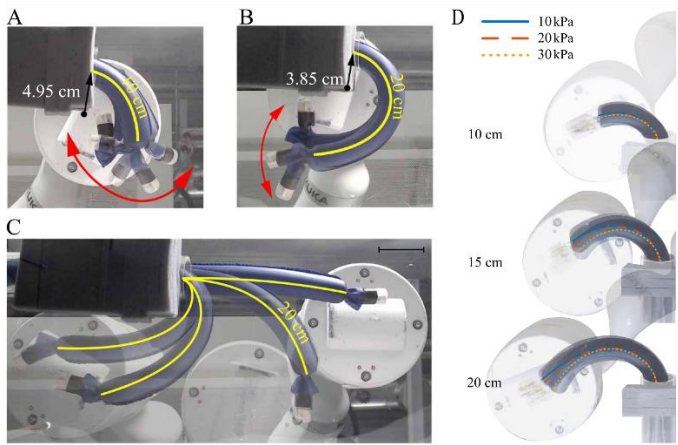

*Figure 6:* *Magnetic steering of the MVR. (A) Minimum bending radius of 4.95 cm with 10 cm of grown body. (B) Minimum bending radius of 3.85 cm with 20 cm of grown body. (C) Demonstration of the ability to buckle the robot body even at a pressure of 30 kPa. (D) Repeatability of achieving bending radii under differing pressures and MVR lengths.*

## Navigation with Suspended Tip

One of the characteristic behaviors of vine robots is the ability to locomote shear-free as a result of their everting nature. The inclusion of a magnetic tip for steering reintroduces a component which moves relative to the environment and therefore would be a source of friction when contact is maintained. To demonstrate the ability to suspend the magnetic tip during growing and steering, the MVR was navigated around a 90-degree bend within a 60 mm diameter perspex tube, illustrated in Figure 7. Here, the EPM was positioned at a height of 160 mm and moved in a fixed pre-planned path in the x-y plane to follow the 90 degree bend. Five repetitions were conducted at two different EPM speeds of 3 mms$^{-1}$ and 4 mms$^{-1}$ under the fixed EPM trajectory. These speeds correspond to just 10% and 20% of the robotic manipulator's maximum speed and were chosen due to limitations in the experimental setup, specifically the manual pressure regulation. Although the robotic manipulator is capable of moving faster, these speeds allowed for the internal pressure to be adjusted, ensuring reliable and controlled navigation. Automating pressure regulation in future work will enable faster speeds to be achieved.

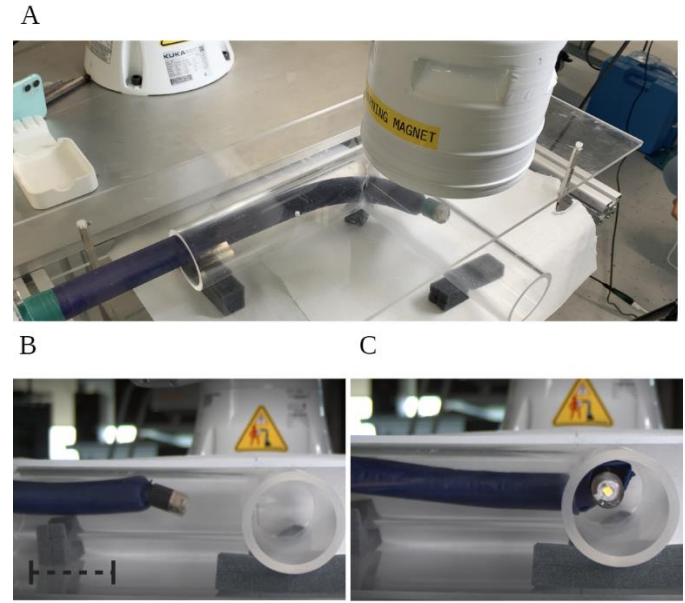

*Figure 7:* *Setup of suspended tip experiment. The robot navigates through a 90-degree turn while keeping the tip suspended, preventing any interaction with the tip and the surrounding tubular environment. (A) Top view. (B) Side view in first half of the 90-degree turn. (C) Side view in the second half of 90-degree turn. Scale Bar: 5 cm.*

Figure 8A shows the 3D position data from the robot's localization system for three independent repetitions at each speed (see supplementary materials for all five repetitions), and the EPM's pre-planned path demonstrating that the robot was repeatedly navigated through the 90-degree path at both speeds. Figure 8B shows the 2D top view of the x-y plane and Figure 8C shows the 2D projected view of the mean and standard deviation of the IPM's center in z (height) versus the distance travelled for each speed. Over five repetitions, an average gap of $11.2 \pm 4.3\ mm$ for 3 mms$^{-1}$ and $12.8 \pm 3.4\ mm$ for 4 mms$^{-1}$ was maintained between the edge of the magnetic tip and the upper surface of the perspex tube (See Supplementary Video 2). EPM-IPM coupling deviations in the x-y plane across all five repetitions for 3 mms$^{-1}$ and 4 mms$^{-1}$ were found to be $19.0 \pm 2.8\ mm$ and $19.6 \pm 1.8\ mm$ in x, $24.3 \pm 3.6\ mm$ and $21.0 \pm 3.3\ mm$ in y and $19.1 \pm 5.8°$ $22.3 \pm 10.3°$ in yaw respectively Deviations for each repetition can be found in Table 1 of the Supplementary Materials. When the EPM is directly above and parallel to the tip magnet, no torque or force is generated in the lateral plane. These EPM-IPM coupling deviations are therefore necessary in order to generate the magnetic wrench required for steering the MVR.

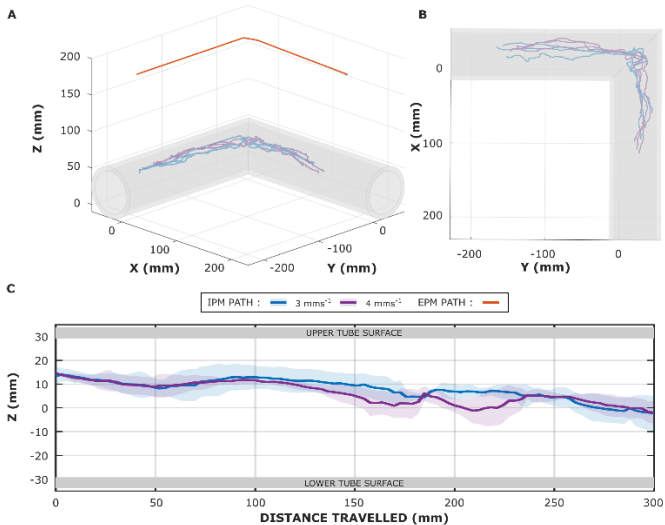

*Figure 8:* Suspended tip experiment data. (A) 3D position data of the center of the EPM and IPM for EPM speeds of 3 mms$^{-1}$ and 4 mms$^{-1}$ for three repetitions within the perspex tube. (B) 2D projected view of the IPM's position on the x-y plane. (C) 2D projected view of the mean and standard deviation of the IPM's z position (height) versus distance travelled along the perspex tube.

## Preliminary Results of Retraction using Magnetic Wrench

In the context of vine robots, retraction refers to the reversal of the growing process, where material inverts at the tip and the vine's length reduces. Without additional constraints on the robot motion however, retraction can fail due to an induced moment from the tensioned tether[26]. To compensate for this, several studies have considered how applying a *grounding force* to the tip can stabilize the vine body, allowing for successful retraction. This force acts in opposition to the tether tension, stabilizing the vine body and therefore allowing the material to be inverted and eventually retracted into the robot base.

If such a force is not applied, the outer material of the robot may collapse first, or in the case of an inflated vine robot, buckling may occur. Buckling during retraction is established as an open problem in vine robots, as it would be desirable for the robot to exit the environment with the same non-shearing behavior as which it entered, particularly in delicate enviroments[27]. Due to vine robots' susceptibility to buckling during this process[27], we propose harnessing magnetic interaction to address this problem. Our focus lies in utilizing magnetic wrench to counteract the buckling phenomenon caused by the pulling of the internal tether from the end of the robot.

To assess the feasibility of this approach, retraction was attempted on a high friction surface under four different conditions. The four scenarios are a deflated MVR with no EPM (Figure 9A), inflated MVR (3 kPa) with no EPM, (Figure 9B), deflated MVR with EPM (Figure 9C) and inflated MVR (3 kPa) with EPM (Figure 9D). In scenario 3 and 4 (Figure 9C&D), the EPM is positioned 60 mm above the magnetic tip and tracked above the tip position. For each retraction attempt, the MVR was first grown in its axial direction to a length of 320 mm and then retracted under manual control.

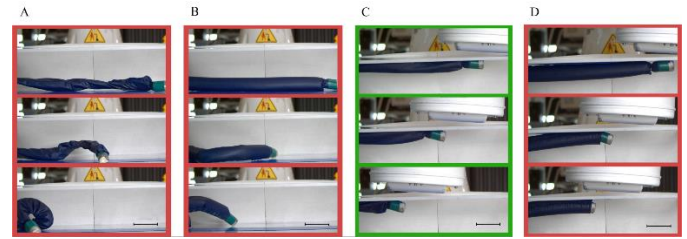

*Figure 9:* Experimental results of retraction under varying conditions. (A) Deflated MVR with no EPM. (B) Inflated MVR with no EPM. (C) Deflated MVR with EPM. (D) Inflated MVR with EPM. Case (C) highlighted green to indicate successful retraction. Scale bar: 5 cm.

In both of the "no EPM" cases (Figure 9A&B), absence of the EPM providing magnetic wrench led to rapid buckling of the body when the internal material was pulled for retraction (See Supplementary Video 3). In contrast, Figure 9C shows that the magnetic wrench generated by the EPM successfully prevented buckling, ensuring the retraction process proceeded smoothly. Interestingly, Figure 9D shows that while buckling did not occur, friction resistance occurred between the material and the tether at the tip, along with the resistance of the sealing ring as a result of the increased pressure, causing the tip to become engulfed occasionally. It is to be noted that, at EPM-IPM separations greater than 60 mm, the magnetic wrench was not sufficient to assist retraction in either case and therefore presentation of this experiment serves only as a preliminary result.

## Demonstration of Navigating a Complex Environment

To evaluate the maneuverability of the MVR, an experiment in which the MVR was navigated using joystick control of the EPM through a complex path of length 0.5 m with consecutive sharp curves (Figure 10A) was conducted. The MVR was navigated through the maze, at which point tip steering was performed before proceeding along a linear trajectory in free space towards a designated target object.

Figure 10B illustrates the ability of the MVR to continue to grow as it successfully navigates through multiple sharp turns. Additionally, tip-concentrated steering of the MVR after negotiating two consecutive sharp turns is shown in the inset of Figure 10B(iii). The inset of Figure 10B(iv) shows the camera's view as the MVR maintains visibility of the target object and direction whilst growing in a linear trajectory, ultimately reaching the target and knocking it off its pedestal. Over five successful repetitions, the average time to navigate the maze was 3 minutes and 15 seconds (See Supplementary Video 4).

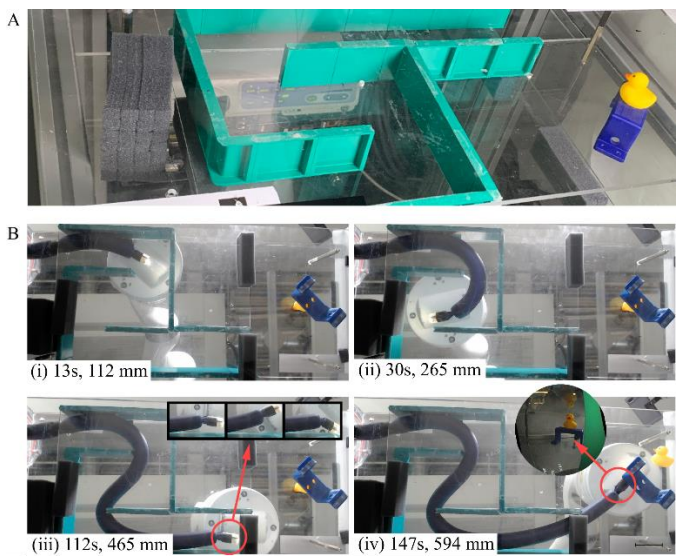

*Figure 10:* Demonstration of MVR navigating the maze environment reporting time and navigation distance. (A) Maze overview. (B) Tiled images of MVR navigating the maze, with the inset in (iii) showing isolated steering of MVR and the inset in (iv) showing the camera view from the MVR tip. Scale Bar: 5 cm.

## DISCUSSION AND CONCLUSION:

In this work, we sought to advance vine robot technology by proposing an external magnetic tip steering mechanism, capable of inducing high curvatures in the MVR body whilst maintaining its compliant, shear-free growing nature. Building on our previous work in magnetic flexible endoscopy, this system forgoes complicated internal mechanisms, instead utilizing a sensorized magnetic tip to facilitate steering, localization and visual feedback. To enable this, we introduced a novel mounting system that not only ensures the secure attachment of the sensorized magnetic element at the MVR's tip during eversion but also provides a reliable method for allowing translation within the inner channel, facilitating seamless transition of sensor cables.

Our investigation of the pushing force of the MVR confirmed the independent influence of both magnetic and pneumatic contributions, with pneumatic contributions being the most significant. Critically, under magnetic manipulation, torque decays with a relationship of $1/r^3$ with the EPM-IPM distance, while force has the relationship $1/r^4$. This relationship highlights the ability of the MVR to be effectively controlled at larger EPM-IPM separation distances (compared to magnetic manipulation alone[21]), relying on magnetic torques for orientation control and pneumatic growing force for forward locomotion. This means that, although downward force is not directly controllable, in the case of MVRs in endoluminal applications, downward force may be achieved through the growing force of the vine while the magnetic torque assists with steering at higher EPM-IPM separation distances. The larger workspace afforded to the system can also seek to accommodate larger patients than would be possible with approaches relying on magnetic manipulation alone[21].

At the design scale evaluated, magnetic steering studies showed that magnetic actuation was found to be dominant over the restoring moment of the pressured section, allowing for high curvature bending of the vine as well as independent tip steering. This exemplifies our design's intent: to focus actuation at the tip while allowing the rest of the vine body to adapt its shape compliantly to the environment; achieving both constant curvature bending when unobstructed, which is dependent on the vine diameter and operating pressure (Figure 6A&B), and non-constant curvatures when bending around an obstruction or a constraint (Figure 6C). This feature is particularly useful in endoluminal scenarios, where adaptability to varying and confined spaces is crucial.

Although the presented MVR requires the addition of a small rigid component at its tip, we have shown that it is possible to suspend the tip during MVR navigation via the balance of magnetic forces and elasticity of the inflated vine body. This approach maintains the shear-free growing motion that is typical of vine robots. It also shows that by virtue of magnetic manipulation, when compared to other mechanisms[15] the weight of the tip is of less concern as it can be compensated for.

In previous work on magnetic robots for colonoscopy, magnetic levitation was used to prevent tissue interaction[28], but it required complex EPM control for gravity compensation and did not address tether drag. Our MVR design reduces harmful tissue interaction and tether drag through friction-free navigation, achieved by the growing vine body and tip suspension, with the tether remaining internal to the outer body. In addition, the soft nature of the growing body could provide a source of stabilization for tooling and for performing procedures such as tissue biopsies.

Addressing retraction challenges[7], our approach of using magnetic wrench was enough to successfully retract the robot in a straight line without buckling the growing body (Figure 9C). However, this method did rely on very close proximity of EPM and IPM and a high friction contact surface. Short EPM-IPM distances limit feasibility for medical applications although larger working distances may be realized by increasing the magnetic volume of the EPM and/or IPM.

The final demonstration of navigating a complex environment (Figure 10) showed that the eversion process is instrumental in facilitating the vine robot's navigation through tortuous environments by eliminating tether drag and enhancing the tip force applied to the magnetic tip. This is an improvement on traditional magnetic endoscopes which struggle to navigate deeper within the anatomy due to friction between the tether and the environment[21]. The maze navigation also highlights the MVR's ability to perform high curvature steering and to continue to grow at arbitrary vine lengths both around obstacles and in free space. In addition, the unique ability to orient the tip independent to the vine body is demonstrated and could be useful to adjust camera and tool orientations in a clinical setting. Although our experiments were conducted in an open-loop manner, our system's design presents a substantial opportunity for closed-loop vine robot control, utilizing our 6 DOF tip localization (shown in Figure 8) and visual servoing. This is supported by the low deviations observed across repetitions, highlighting the stability of the MVR behavior.

Compared to other vine robot steering mechanisms, such as tendon-driven or pneumatic muscles, our design enables high curvature, tip focused steering, and the unique ability to steer the tip independently of the vine body. The associated design simplicity, devoid of internal motorized components and rigid tip mounts, can enable smaller scales and inherent safety, which is conducive to medical applications like endoluminal navigation (e.g. gastrointestinal endoscopy, bronchoscopy or cardiovascular navigation), but is limited to scenarios amenable to external magnetic field generation. MVRs also retain the benefits of shear-free motion due to the combination of everting "growth" motion, which simply adds material at the tip to move forward, with the magnetic suspension of a short, rigid component at the tip that is smaller in diameter than the outer vine.

The collective features of MVRs can thus minimize the risk of complications for the patient associated with friction-based movement, offering a gentle and safe option for navigation within delicate biological structures. Additionally, the higher tip forces generated by the pneumatic component and elimination of tether drag may allow MVR's to navigate deeper into endoluminal anatomy than conventional magnetic endoscopes[21] or catheters[23].

Although we have successfully demonstrated the feasibility of magnetic manipulation for vine robots, the general scalability of MVRs remains to be investigated. Magnetic manipulation systems have been successfully demonstrated at smaller scales, however with miniaturized MVRs with lower overall magnetic volume, it remains to be proven if steering of the pressurized body can still be achieved. A systematic design study would be required to evaluate the feasibility of the concept across multiple scales, as our current design was motivated by endoluminal applications (specifically for gastrointestinal endoscopy), where a camera, localization and tool channel are required.

Overall, our findings highlight the first successful integration of magnetic steering mechanisms into vine robots, expanding their dexterity while maintaining shear-free navigation for endoluminal applications. Future research will be directed towards enhancing the MVR's performance by implementing closed-loop control for visual servoing, tip suspension and retraction in complex environments. We aim to further validate our technology's practical utility in clinically relevant experiments, incorporating a larger magnetic volume to extend the EPM-IPM separation distance and improve retraction efficiency. We also aim to investigate the miniaturization potential for use in smaller lumens to extend our technology to a wider range of endoluminal applications.

## ACKNOWLEDGEMENTS

Research reported in this article was supported in part by the Korea Health Technology R&D Project through the Korea Health Industry Development Institute (KHIDI), funded by the Ministry of Health & Welfare, Republic of Korea (grant number: HI22C079600), the NAVER Digital Bio Innovation Research Fund, funded by NAVER Corporation (Grant No. 3720230130), the European Research Council (ERC) under grant agreement no. 818045 (NoLiMiTs), the European Union's Horizon 2020 Research and Innovation Programme under grant agreement no. 952118 (AUTOCAPSULE), by the Engineering and Physical Sciences Research Council (EPSRC) under grants #EP/R045291/1, #EP/V047914/1, #EP/V009818/1, by the National Institute of Biomedical Imaging and Bioengineering of the National Institutes of Health (NIH) under Award Number R01EB018992 and by the National Institute for Health and Care Research (NIHR) Leeds Biomedical Research Centre. Any opinions, findings, conclusions, or recommendations expressed in this article are those of the authors and do not necessarily reflect the views of the KHIDI, NAVER, ERC, EPSRC, NIH, NHS, or the NIHR.

We would like to thank James William Martin, Robyn Fynn and Samwise Wilson for their technical support throughout this project. Figures 1 and 2 were created with Biorender.com.

## AUTHOR CONTRIBUTIONS STATEMENT

*Nam Gyun Kim and Nikita Greenidge contributed equally to the work and are to be considered co-first authors on this publication.

Nam Gyun Kim: Design and development of mechatronic systems (lead), writing and editing (supporting). Nikita Greenidge: Design of magnetic control systems (lead), writing, reviewing and editing (equal). Joshua Davy: Design of magnetic control systems (supporting), writing, reviewing and editing (equal). Shinwoo Park: Design and development of mechatronic systems (supporting). James H. Chandler: Reviewing and editing (supporting). Jee-Hwan Ryu: Reviewing and editing (supporting). Pietro Valdastri: Reviewing and editing (lead)

## AUTHOR DISCLOSURE STATEMENT

No competing financial interests exist.

## BIBLIOGRAPHY

1.  Mishima D, Aoki T, Hirose S. Development of pneumatically controlled expandable arm for search in the environment with tight access. Field and Service Robotics: Recent Advances in Reserch and Applications. 2006:509-18.

2.  Hawkes EW, Blumenschein LH, Greer JD, Okamura AM. A soft robot that navigates its environment through growth. Science Robotics. 2017 Jul 19;2(8):eaan3028.

3.  Berthet-Rayne P, Sadati SH, Petrou G, Patel N, Giannarou S, Leff DR, Bergeles C. Mammobot: A miniature steerable soft growing robot for early breast cancer detection. IEEE Robotics and Automation Letters. 2021 Mar 26;6(3):5056-63.

4.  Li M, Obregon R, Heit JJ, Norbash A, Hawkes EW, Morimoto TK. Vine catheter for endovascular surgery. IEEE Transactions on Medical Robotics and Bionics. 2021 Mar 31;3(2):384-91.

5.  Valdastri P, Simi M, Webster III RJ. Advanced technologies for gastrointestinal endoscopy. Annual review of biomedical engineering. 2012 Aug 15;14:397-429.

6.  Luong J, Glick P, Ong A, DeVries MS, Sandin S, Hawkes EW, Tolley MT. Eversion and retraction of a soft robot towards the exploration of coral reefs. In2019 2nd IEEE International Conference on Soft Robotics (RoboSoft) 2019 Apr 14 (pp. 801-807). IEEE.

7.  Coad MM, Blumenschein LH, Cutler S, et al. Vine robots: Design, teleoperation, and deployment for navigation and exploration. *IEEE Robotics & Automation Magazine*. 2019;27(3):120-132.

8.  Greer JD, Morimoto TK, Okamura AM, Hawkes EW. Series pneumatic artificial muscles (sPAMs) and application to a soft continuum robot. In2017 IEEE International Conference on Robotics and Automation (ICRA) 2017 May 29 (pp. 5503-5510). IEEE.

9. Greer JD, Morimoto TK, Okamura AM, Hawkes EW. A soft, steerable continuum robot that grows via tip extension. Soft robotics. 2019 Feb 1;6(1):95-108.

10. Naclerio ND, Hawkes EW. Simple, low-hysteresis, foldable, fabric pneumatic artificial muscle. IEEE Robotics and Automation Letters. 2020 Feb 26;5(2):3406-13.

11. Naclerio ND, Karsai A, Murray-Cooper M, et al. Controlling subterranean forces enables a fast, steerable, burrowing soft robot. *Science Robotics*. 2021;6(55).

12. Maur PA der, Djambazi B, Haberthür Y, et al. Roboa: Construction and evaluation of a steerable vine robot for search and rescue applications. In: *2021 IEEE 4th International Conference on Soft Robotics (RoboSoft)*. IEEE; 2021:15-20.

13. Haggerty DA, Naclerio ND, Hawkes EW. Hybrid vine robot with internal steering-reeling mechanism enhances system-level capabilities. *IEEE Robotics and Automation Letters*. 2021;6(3):5437-5444.

14. Lee DG, Kim NG, Ryu JH. High-curvature consecutive tip steering of a soft growing robot for improved target reachability. In: *2023 IEEE/RSJ International Conference on Intelligent Robots and Systems (IROS)*. IEEE/RSJ; 2023:6477-6483.

15. Jeong SG, Coad MM, Blumenschein LH, et al. A tip mount for transporting sensors and tools using soft growing robots. In: 2020 IEEE/RSJ International Conference on Intelligent Robots and Systems (IROS). IEEE/RSJ; 2020:8781-8788.

16. Heap WE, Naclerio ND, Coad MM, et al. Soft retraction device and internal camera mount for everting vine robots. In: 2021 IEEE/RSJ International Conference on Intelligent Robots and Systems (IROS). IEEE/RSJ; 2021:4982-4988.

17. Kim J, Jang J, Lee S, Jeong SG, Kim YJ, Ryu JH. Origami-inspired new material feeding mechanism for soft growing robots to keep the camera stay at the tip by securing its path. *IEEE Robotics and Automation Letters*. 2021;6(3):4592-4599.

18. Norton JC, Slawinski PR, Lay HS, Martin JW, Cox BF, Cummins G, Desmulliez MP, Clutton RE, Obstein KL, Cochran S, Valdastri P. Intelligent magnetic manipulation for gastrointestinal ultrasound. Science robotics. 2019 Jun 19;4(31):eaav7725.

19. Kim Y, Parada GA, Liu S, Zhao X. Ferromagnetic soft continuum robots. Science Robotics. 2019 Aug 28;4(33):eaax7329.

20. Pittiglio G, Chandler JH, Veiga T da, et al. Personalized magnetic tentacles for targeted photothermal cancer therapy in peripheral lungs. *Communications Engineering*. 2023;2(1):50.

21. Martin JW, Scaglioni B, Norton JC, et al. Enabling the future of colonoscopy with intelligent and autonomous magnetic manipulation. *Nature Machine Intelligence*. 2020;2(10):595-606.

22. Gervasoni S, Lussi J, Viviani S, Boehler Q, Ochsenbein N, Moehrlen U, Nelson BJ. *Magnetically assisted robotic fetal surgery for the treatment of spina bifida. IEEE Transactions on Medical Robotics and Bionics.* 2022 Feb 3;4(1):85-93.

23. Pittiglio G, Brockdorff M, Veiga T da, Davy J, Chandler JH, Valdastri P. Collaborative magnetic manipulation via two robotically actuated permanent magnets. *IEEE Transactions on Robotics*. Published online 2022.

24. Taddese AZ, Slawinski PR, Pirotta M, De Momi E, Obstein KL, Valdastri P. Enhanced real-time pose estimation for closed-loop robotic manipulation of magnetically actuated capsule endoscopes. The International journal of robotics research. 2018 Jul;37(8):890-911.

25. Blumenschein LH, Okamura AM, Hawkes EW. Modeling of bioinspired apical extension in a soft robot. In: *Biomimetic and Biohybrid Systems: 6th International Conference, Living Machines 2017, Stanford, CA, USA, July 26–28, 2017, Proceedings 6*. Springer; 2017:522-531.

26. Kim, N.G., Seo, D., Park, S. and Ryu, J.H., 2023. Self-Retractable Soft Growing Robots for Reliable and Fast Retraction While Preserving Their Inherent Advantages. *IEEE Robotics and Automation Letters*

27. Coad MM, Thomasson RP, Blumenschein LH, Usevitch NS, Hawkes EW, Okamura AM. Retraction of soft growing robots without buckling. *IEEE Robotics and Automation Letters*. 2020;5(2):2115-2122.

28. Pittiglio G, Barducci L, Martin JW, Norton JC, Avizzano CA, Obstein KL, Valdastri P. Magnetic levitation for soft-tethered capsule colonoscopy actuated with a single permanent magnet: A dynamic control approach. IEEE robotics and automation letters. 2019 Jan 23;4(2):1224-31.

# TABLES

Table 1: Mean bending radii of the MVR's final position across a range of robot lengths and pressures under fixed circular EPM trajectories.

| ROBOT LENGTH (mm) | PRESSURE (kPa) | MEAN BENDING RADIUS (mm) | BENDING RADIUS STANDARD DEVIATION (mm) |
|---|---|---|---|
| 100 | 10.0 | 65.5 | 0.3 |
| 100 | 20.0 | 66.3 | 0.2 |
| 100 | 30.0 | 67.0 | 0.1 |
| 150 | 10.0 | 67.3 | 0.5 |
| 150 | 20.0 | 69.7 | 1.1 |
| 150 | 30.0 | 69.0 | 0.4 |
| 200 | 10.0 | 63.4 | 1.5 |
| 200 | 20.0 | 72.2 | 0.4 |
| 200 | 30.0 | 66.5 | 0.9 |